\documentclass{article}

\usepackage{graphicx}
\usepackage{amsmath}
\usepackage{comment}
\usepackage{tcolorbox}
\usepackage{subfigure}
\usepackage[makeroom]{cancel}

\usepackage[preprint]{neurips_2025}

\usepackage[utf8]{inputenc} 
\usepackage[T1]{fontenc}    
\usepackage{hyperref}       
\usepackage{url}            
\usepackage{booktabs}       
\usepackage{amsfonts}       
\usepackage{nicefrac}       
\usepackage{microtype}      
\usepackage{xcolor}         

\newcommand{\zm}[1]{{\color{black!0!blue} #1}}

\title{Neural Thermodynamic Laws for Large Language Model Training}

\author{%
  Ziming Liu\thanks{zmliu@mit.edu},\  Yizhou Liu,\ Jeff Gore,\ Max Tegmark \\
  Massachusetts Institute of Technology \\
}

\begin{document}

\maketitle

\begin{abstract}
  Beyond neural scaling laws, little is known about the laws underlying large language models (LLMs). We introduce \textit{neural thermodynamic laws} (NTL) -- a new framework that offers fresh insights into LLM training dynamics.  On the theoretical side, we demonstrate that key thermodynamic quantities (e.g., temperature, entropy, heat capacity, thermal conduction) and classical thermodynamic principles (e.g., the three laws of thermodynamics and the equipartition theorem) naturally emerge under river-valley loss landscape assumptions. On the practical side, this scientific perspective yields intuitive guidelines for designing learning rate schedules.
\end{abstract}

\begin{figure}[htbp]
    \centering
    \includegraphics[width=0.94\linewidth]{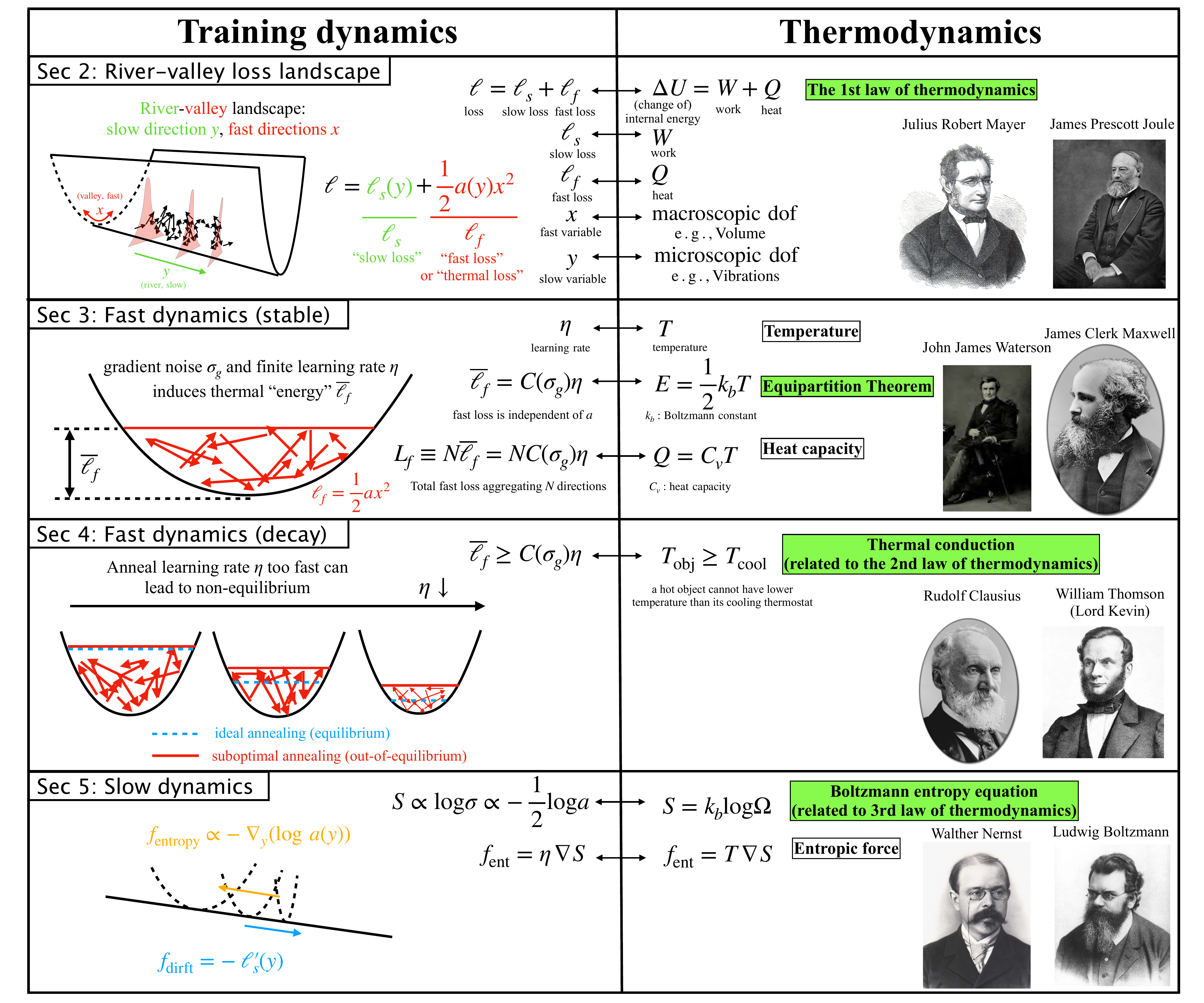}
    \caption{Connections between LLM training dynamics and thermodynamics.}
    \label{fig:training-thermo-duality}
\end{figure}

\section{Introduction}

Large neural networks bear striking similarities to thermodynamic systems -- both involve a vast number of degrees of freedom and exhibit stochastic dynamics. It is therefore not surprising that connections between neural networks and thermodynamics have been explored in prior work~\cite{welling2011bayesian, mehdi2024thermodynamics,engel2001statistical,bahri2020statistical}. However, these studies primarily focus on classical machine learning models with relatively well-understood loss landscapes. In contrast, recent research has only begun to shed light on the complex loss landscapes of large language models (LLMs), characterized by the so-called ``river-valley" structure -- comprising sharp, fast directions (valley) and flat, slow directions (rivers)~\cite{wen2024understanding,wei2019noise,liu2025focus}. Intuitively, the fast dynamics rapidly ``equilibrate'' within valleys, while the slow dynamics evolve gradually along rivers, subtly modulated by the fast components. The goal of this paper is to formalize this intuition through the lens of \textit{Neural Thermodynamic Laws} (NTL). We show that key thermodynamic quantities and principles -- including temperature, entropy, heat capacity, thermal conduction, the three laws of thermodynamics, and the equipartition theorem -- emerge naturally from the training dynamics of LLM (see the connections between training dynamics and thermodynamics in Figure~\ref{fig:training-thermo-duality}).

The duality between LLM training dynamics and thermodynamics is not only conceptually and theoretically compelling, but also provides practical insights -- for example, into the design of learning rate schedules. A common learning rate schedule used in LLM pretraining is the warmup-stable-decay (WSD). According to~\cite{wen2024understanding,hu2024minicpm}, the stable phase corresponds to motion along the river, with fluctuations in the valley directions, while the decay phase suppresses these valley variations. Motivated by this, we introduce a toy model of the river-valley landscape. This model is analytically solvable, admits a natural thermodynamic interpretation, and shows strong empirical agreement with actual LLM training dynamics.

The paper is organized as follows. The timescale separation between fast and slow dynamics allows us to decompose the total loss function $\ell$ into two components -- the fast part $\ell_f$ and the slow part $\ell_s$, which motivates our toy model of the river-valley landscape (Section~\ref{sec:2}). With a fixed learning rate, the fast dynamics converge to a steady-state distribution, analogous to thermal equilibrium (Section~\ref{sec:3}). When the learning rate decays, the distribution evolves accordingly -- resembling annealing (Section~\ref{sec:4}). Moreover, the fast dynamics exert an effective entropic force on the slow dynamics, similar to entropic forces in physics (Section~\ref{sec:5}). Notably, the learning rate $\eta$ plays a central role in all of these phenomena. By clarifying its complex and sometimes contradictory effects, we derive an intuitive guideline for designing efficient learning rate schedules (Section~\ref{sec:6}), followed by related works (Section~\ref{sec:related_works}) and conclusions (Section~\ref{sec:conclusions}).

%


Unlike prior work that approaches LLM optimization, especially the design of learning rate schedules, from largely empirical or phenomenological perspectives, our characterizations are more mechanistic. Our technical contributions are as follows:

\begin{itemize}

    \item {\bf Formulation of Fast-slow decomposition}. In river-valley landscapes, we decompose training into two processes: (1) fast dynamics: either \textit{equilibrium} (under fixed $\eta$) or \textit{annealing} (under decaying $\eta$) along the \textit{valley} and (2) slow dynamics: \textit{drift} along the \textit{river}.
    \item {\bf An exactly solvable toy model.} We introduce a tractable toy model of the river-valley landscape that captures both fast and flow dynamics. This model admits analytical solutions for training behavior and optimal learning rate schedules. 
    \item {\bf Empirical relevance to LLMs.} We demonstrate that insights  from the toy model generalize well to real LLM training, providing intuitive and effective heuristics for the learning rate schedule.
    \item {\bf A bridge to physics} The duality between neural network training and thermodynamics provides a foundation for developing a deeper scientific understanding of deep learning.
\end{itemize}

\section{River Valley Loss Landscape}\label{sec:2}


Recent work~\cite{wen2024understanding} showed that LLM loss landscape resembles a river-valley landscape: a flat river lies at the bottom of sharp valleys. Training slowly progresses along the river while bouncing quickly between the sharp hillsides. Throughout the paper, we interchangeably use valley dynamics = fast dynamics, river dynamics = slow dynamics. 

{\bf A dilemma for learning rate $\eta$} A good learning rate should strike a good balance between the two objectives: (A) enabling progress along the river directions -- where the loss typically decreases monotonically -- which favors a large $\eta$; and (B) minimizing variance along the valley directions -- which favors small $\eta$. In WSD schedules, the stable phase takes care of (A), while the decay phase takes care of (B)~\cite{wen2024understanding}. To better understand this trade-off, we introduce a toy model that admits analytical characterization of the training dynamics along both river and valley directions.

\subsection{Toy model}
The toy model $\ell(x,y)=c(y) + \frac{1}{2}a(y)x^2$ is a loss function in 2D, resembling the river-valley landscape in the top left of Figure~\ref{fig:training-thermo-duality}. It consists of a fast variable $x$ and a slow variable $y$. For any fixed $y$, the loss is minimized at $x=0$, which traces out the riverbed at the bottom of the valley, with corresponding loss $c(y)$. The loss function decomposes additively into two components: the valley component $\ell_f(x,y)\equiv \frac{1}{2}a(y)x^2$ (called \textit{fast loss} or \textit{thermal loss}) and the river component $\ell_s(x,y)\equiv c(y)$ (called \textit{slow loss}). In the remainder of the paper, we analyze the training dynamics of SGD and SignGD on this landscape -- under learning rate $\eta$ and gradient noise $\sigma_g$ -- and demonstrate its relevance to the training behavior of large language models.


\subsection{Thermodynamics: First law of thermodynamics}
The decomposition of fast and slow dynamics is reminiscent of quasi-static equilibrium in thermodynamics. Consider a piston slowly changing the volume of a gas-filled chamber: while the piston (a slow variable) moves slowly, the gas molecules (fast variables) undergo rapid thermal motion and quickly reach a new thermal equilibrium. According to the first law of thermodynamics $\Delta U = W + Q$, the change in internal energy $\Delta U$ is composed of work $W$ (associated with slow dynamics) and heat $Q$ (associated with fast dynamics). This mirrors the decomposition of the loss function in the river-valley landscape $\ell=\ell_s + \ell_f$, where $\ell_s$ captures slow dynamics and $\ell_f$ captures fast dynamics. In this analogy, the slow variable $y$ corresponds to macroscopic quantities such as volume, while the fast variable $x$ corresponds to microscopic degrees of freedom, such as atomic vibrations. In the next section, we examine how the fast fluctuations in $x$ contribute to the fast loss component $\ell_f$. Although this decomposition is conceptually interesting, it does not yet provide a quantitative characterization of $\ell_f$ and $\ell_s$, which we aim to address in the remainder of the paper.

\section{Valley Dynamics in Equilibrium (Stable phase)}\label{sec:3}

The fast-slow separation allows us to treat the slow variable $y$ as fixed while analyzing the fast dynamics. Relevant to the fast dynamics is the fast loss $\ell_f(x,y)\equiv \frac{1}{2}a(y)x^2$. For simplicity, we drop the dependence on $y$ and write $\ell_f(x)=\frac{1}{2}ax^2$. We consider stochastic gradient descent (SGD) or signed gradient descent (SignGD)~\footnote{For simplicity, we set momentums to zero. Note that SignGD is a special case of Adam when $(\beta_1,\beta_2)=(0,0)$.} on this quadratic function, with learning rate $\eta$ and gradient noise $\sigma_g$. Under this quadratic approximation, the training dynamics converge to a Gaussian steady-state distribution, $p(x)=\frac{1}{\sqrt{2\pi}\sigma}{\rm exp}(-\frac{x^2}{2\sigma^2})$, characterized by width $\sigma$. Our goal is to understand how $\sigma$ depends on sharpness $a$, learning rate $\eta$, and gradient noise $\sigma_g$. In Section~\ref{subsec:valley_eq_toy}, we derive the functional form $\sigma = \sigma(\eta, a,\sigma_g)$ for SGD and SignSGD. Section~\ref{subsec:valley_eq_thermo} provides a thermodynamic interpretation of the results, followed by empirical validation on LLM training in Section~\ref{subsec:valley_eq_exp}.

\subsection{Toy model: steady distribution}\label{subsec:valley_eq_toy}

{\bf SGD} An SGD optimizer with learning rate $\eta$ and gradient noise $\sigma_g$ obeys the following dynamics:
\begin{equation}\label{eq:sgd}
    x_{t+1} = x_t - \eta(a x_t + \sigma_g\dot{W}),
\end{equation}
where $\dot{W}\sim \mathcal{N}(0,1)$ is the standard Brownian motion. The equilibrium distribution is the Gaussian distribution $p_\sigma(x)=\frac{1}{\sqrt{2\pi}\sigma}e^{-\frac{x^2}{2\sigma^2}}$ (see derivtions in Appendix~\ref{app:sgd-converge-gaussian}). In equilibrium, the variance is preserved, i.e., ${\rm Var}(x_{t+1})={\rm Var}(x_t)$, giving rise to the equation $\sigma^2 = (1-\eta a)^2\sigma^2 + \sigma_g^2$, resulting in $\sigma = \sigma_g/\sqrt{a(2/\eta-a)}$. This formula is only well-defined when $0<a<\frac{2}{\eta}$. When $a\to 2/\eta$, the learning rate reaches the so-called ``edge of stability''~\cite{cohen2021gradient}. Since most directions in an over-parametrized model are relatively flat, we are interested in the flat limit when $a\ll 2/\eta$, which simplifies $\sigma$ to $\sigma\approx \sqrt{\eta/(2a)}\sigma_g\propto \eta^{1/2}a^{-1/2}\sigma_g^1$. 

{\bf SignGD} An SignGD optimizer with learning rate $\eta$ and gradient noise $\sigma_g$ obeys the dynamics:
\begin{equation}\label{eq:signgd}
    x_{t+1} = x_t - \eta\ {\rm sign}(ax_t+\sigma_g\dot{W}).
\end{equation}
Using the variance preservation condition as in SGD, we derive the steady-state Gaussian width $\sigma = \frac{\sqrt{\pi}}{4}\eta \sqrt{1+\sqrt{1+\frac{32}{\pi}(\frac{\sigma_g}{a\eta})^2}}$. The flat limit $a\ll \sigma_g/\eta$ gives $\sigma\approx (\frac{\pi}{8})^{1/4}\sqrt\frac{\sigma_g\eta}{a}\propto \eta^{1/2}a^{-1/2}\sigma_g^{1/2}$. Detailed derivations are postponed to Appendix~\ref{app:signgd}. Results are summarized in Table~\ref{tab:1D_toy} for reference.
\begin{table}[]
    \centering
    \caption{Optimization on the 1D quadratic function $\ell(x)=\frac{1}{2}ax^2$}
    \begin{tabular}
    {|c|c|c|}\hline
    Optimizer & SGD & SignGD  \\\hline
    Equation  & $x_{t+1}=x_t - \eta(a x_t+\sigma_g\dot{W})$ & $x_{t+1}=x_t - \eta\ {\rm sign}(a x_t+\sigma_g\dot{W})$ \\\hline
    Steady distribution width $\sigma$ & $\frac{\sigma_g}{\sqrt{a(\frac{2}{\eta}-a)}}$ & $\frac{\sqrt{\pi}}{4}\eta\sqrt{1+\sqrt{1+\frac{32}{\pi}(\frac{\sigma_g}{a\eta})^2}}$ \\\hline
    $\sigma$ (flat limit $a\eta\ll 1$) & $\sqrt{\frac{\eta}{2a}}\sigma_g\propto \eta^{1/2}a^{-1/2}\sigma_g$ & $(\frac{\pi}{8})^{1/4}\sqrt{\frac{\sigma_g\eta}{a}}\propto\eta^{1/2}a^{-1/2}\sigma_g^{1/2}$ \\\hline
    Thermal loss $\overline{\ell_f}$ & $\frac{\sigma_g^2}{4}\eta$ & $\sqrt{\frac{\pi}{32}}\sigma_g\eta$ \\\hline
    Optimal decay schedule & $\eta_t = \frac{\frac{\eta}{2}}{1+\frac{t}{t_h}}\ (t_h=\frac{2}{a\eta})$ & $\eta_t = \frac{\frac{\eta}{2}}{1+\frac{t}{t_h}}\ (t_h=\sqrt{2\pi}\frac{\sigma_g}{a\eta})$ \\\hline
    \end{tabular}
    \label{tab:1D_toy}
\end{table}

{\bf Thermal loss} The averaged thermal loss is $\overline{\ell_f} = \mathbb{E}_{x\sim p_\sigma(x)}(\frac{1}{2}ax^2)=\frac{1}{2}a\sigma^2$. Notice that for both SGD and SignGD, $\sigma\propto a^{-1/2}$, it follows that $a$ is cancelled out in $\overline{\ell_f}$, i.e., $\overline{\ell_f}\propto \eta\sigma_g^2$ for SGD and $\ell_f\propto \eta\sigma_g$ for SignGD. The independence of $a$ leads to an interesting fact: given two directions with different sharpness $a_1\neq a_2$, they induce the same $\overline{\ell_f}$ (as long as $\eta$ and $\sigma_g$ are the same). This also means: no matter how long the stable phase goes on (resulting in different points along the river with different sharpnesses), the reducible thermal loss is roughly the same in the decay phase, which explains the observations in~\cite{hagele2024scaling}. The averaged thermal loss $\overline{\ell_f}$ can also be interpreted as being averaged over many valley directions.

\subsection{Thermodynamics: Equi-partition theorem, Temperature, Heat capacity}\label{subsec:valley_eq_thermo}
The independence of sharpness corresponds to the \textit{equipartition theorem} in thermodynamics, which states that: in a system in thermal equilibrium, energy is distributed equally among all degrees of freedom that appear quadratically in the system's energy. Quantititavely, $E=\frac{1}{2}k_bT$, Where $k_b$ is the Boltzmann constant and $T$ is the temperature. In particular, the energy of a vibrational degree of freedom is independent of the spring constant, which corresponds to sharpness $a$ in our case. Now we can make a mapping between optimization $\overline{\ell_f}\propto \sigma_g^n\eta$ ($n=1$ for SignGD, $n=2$ for SGD) and thermodynamics $E=\frac{1}{2}k_bT$. Ignoring constants, we have effective temperature $T\sim \eta$, i.e., the learning rate $\eta$ can be interpreted as \textit{temperature}. Given this, the slope $C\equiv\frac{\partial \overline{\ell_f}}{\partial \eta}$ can be interpreted as \textit{heat capacity}. Now we can simply relate $\overline{\ell_f}$ and $\eta$ by $\overline{\ell_f}=C\eta$. When there are $N$ valley directions, the total thermal loss is summed over all valley directions $\overline{L_f}=N\overline{\ell_f}=NC\eta$.

\subsection{Experiments}\label{subsec:valley_eq_exp}

\begin{figure}
    \centering
    \includegraphics[width=1.0\linewidth]{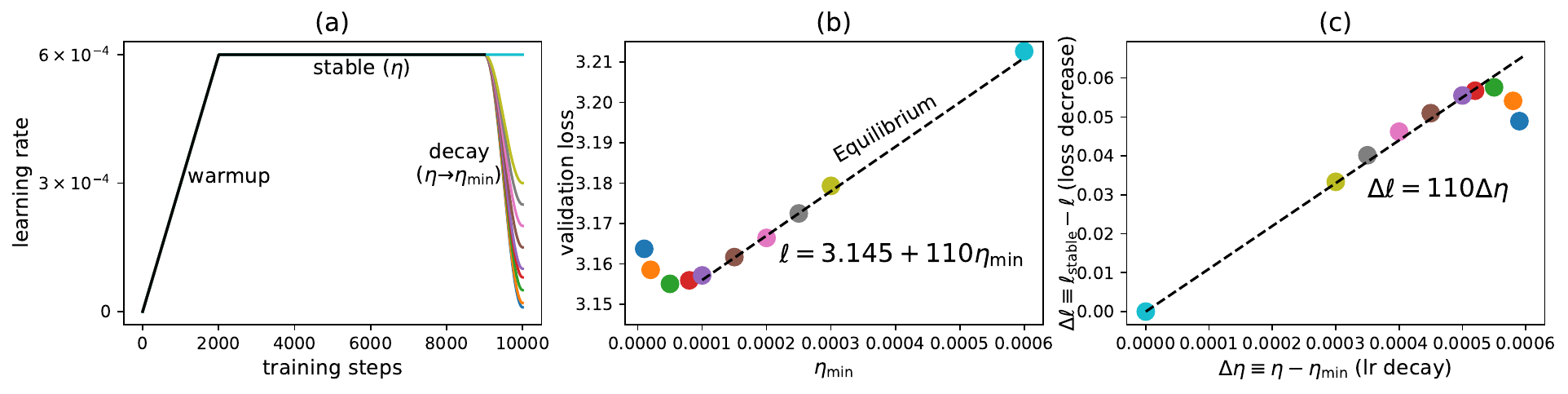}
    \caption{(a) LLM pretraining usually uses the WSD (warmup-stable-decay) learning rate schedule. $\eta_{\rm min}$ is the final learning rate. (b) validation loss is a linear function of $\eta_{\rm min}$ for large $\eta_{\rm min}$. (c) $\Delta\ell$ is a linear function of $\Delta\eta$ for small $\Delta\eta$.}
    \label{fig:eta_min}
\end{figure}

{\bf GPT-2 Experiment setup:} We pre-train a GPT-2-small model (based on NanoGPT~\cite{Karpathy2022}) on OpenWebText. We use 8 V100 GPUs,
choose block size 1024, batch size 480 blocks. We use the Adam Optimizer, with warmup-stable-decay learning rate schedules as shown in Figure~\ref{fig:eta_min} (a). We always use 2000-step linear warmup from 0 to $6\times 10^{-4}$. The stable phase has a learning rate $\eta$. The decay phase starts from $\eta$ and cosine decays to $\eta_{\rm min}$. The total number of training steps is 10k.

We have shown in the toy model that the averaged thermal loss $\overline{\ell_f}$ is linear to the learning rate $\eta$ (assuming thermal equilibrium). We now show that this relation holds for large language models. We have 2k-step warmup, 7k-step stable ($\eta=6\times 10^{-4}$) and 1k-step cosine decay to $\eta_{\rm min}$, which is swept. In Figure~\ref{fig:eta_min} (b), we show that the final validation loss is linear to $\eta_{\rm min}$ for large $\eta_{\rm min}$. Since the decay phase is short, we can assume that $\ell_s$ does not vary much across decay schedules. As a result, $\ell$ is representative of $\overline{\ell_f}$, and we measure that $\ell=3.145+110\eta_{\rm min}$. Comparing to the theoretical thermal loss $\overline{L_f}=(\sqrt{\frac{\pi}{32}}N\sigma_g)\eta$ (for SignGD~\footnote{SignGD is a special cases of Adam when $(\beta_1,\beta_2)=(0,0)$.}), we have $N=\frac{110}{\sigma_g}\sqrt{\frac{32}{\pi}}\approx 5\times 10^6=5{\rm M}$ where $\sigma_g\approx 7\times 10^{-5} $ is estimated using batches. Note that GPT-2 small has 124M parameters, and 5M valley directions are only 4\% of its total parameters. This is expected: most directions of an over-parameterized model are flat (``river''), and only a small portion of directions are sharp (``valley''). However, too small $\eta_{\rm min}$ leads to higher loss, and the lowest loss occurs around $\eta_{\rm min}\approx 5\times 10^{-5} > 0$. The non-monotonic behavior at small $\eta_{\rm min}$ is due to the breakdown of thermal equilibrium when $\eta$ decays too fast, as we will elaborate on in the next section. 

\section{Valley Dynamics in Annealing (Decay phase)}\label{sec:4}

In the last section, we have shown that the width of the steady distribution $\sigma$ depends on the learning rate $\eta$ as $\sigma\propto\sqrt{\eta}$. In the decay phase when $\eta$ decays, $\sigma$ will decay as a result. If $\eta$ decays slowly enough, we could expect that $\sigma\propto\sqrt{\eta}$ always holds because of quasi-static thermal equilibrium. But is this optimal? Probably no. But the other extreme -- too fast $\eta$ decay -- is also not optimal. This dilemma is because $\eta$ plays two roles: (1) $\eta$ is temperature that controls the Gaussian noise (we want $\eta$ to be small); (2) $\eta$ is step size that controls the time scale (we want $\eta$ to be large). There exists an  ``optimal'' $\eta$ decay schedule in the sense that $\ell_f$ is reduced as quickly as possible.

\subsection{Toy model: optimal learning rate decay}

Now we consider a learning rate decay schedule, i.e., a sequence $\eta_0\geq \eta_1\geq\eta_2\geq \cdots\geq \eta_T$. the dynamics of SGD (Eq.~(\ref{eq:sgd})) now becomes $x_{t+1} = x_t - \eta_t(a x_t + \sigma_g\dot{W})$. Since the equation is linear, if $p(x_0)$ starts off as a Gaussian distribution, $p(x_t)$ remains a Gaussian distribution (with time-varying Gaussian width, denoted as $\sigma_t$) for all $t\geq 0$.
Assuming that at $t=0$, $p(x_0)$ is in thermal equilibrium with learning rate $\eta$ whose Gaussian width is $\sigma_0$. Gaussian widths $\sigma_t$ obeys the following recursive relation $\sigma_{t+1}^2 = (1-\eta_t a)^2\sigma_t^2 + (\eta_t\sigma_g)^2 = (a^2\sigma_t^2+\sigma_g^2)\eta_t^2 - 2a\sigma_t^2\eta_t + \sigma_t^2$, 
which is a quadratic function of $\eta_t$. We want to choose $\eta_t$ such that $\sigma_{t+1}$ is minimized $\eta_t = \frac{a\sigma_t^2}{a^2\sigma_t^2+\sigma_g^2},\ \sigma_{t+1}^2 = \frac{\sigma_g^2\sigma_t^2}{\sigma_g^2+a^2\sigma_t^2}$. 
By inverting the second equation, we get $\frac{1}{\sigma_{t+1}^2} = \frac{1}{\sigma_t^2} + \frac{a^2}{\sigma_g^2}$,
which means that $\{\frac{1}{\sigma_t^2}\}$ forms an arithmetic sequence. It is clear that $\frac{1}{\sigma_t^2}=\frac{1}{\sigma_0^2} + \frac{a^2t}{\sigma_g^2}$. Correspondingly,
\begin{equation}\label{eq:optimal_schedule}
    \eta_t =  \frac{\frac{1}{a}}{1+\frac{\sigma_g^2}{a^2}(\frac{1}{\sigma_0^2}+\frac{a^2t}{\sigma_g^2})} \approx \frac{\frac{\eta}{2}}{1+\frac{t}{t_h}}\ (t_h\equiv \frac{2}{a\eta}),
\end{equation}
where $t_h$ is the time needed to decrease $\eta_t$ by half, hence representing the characteristic time scale of learning rate decay. Recall that $\overline{\ell_f}=C\eta$, so $\overline{\ell_{f,t}}=C\eta_t$ has the same decay form as $\eta_t$. We make a few remarks about the optimal schedule.

{\bf Remark 1: Asymptotic behavior} As $t\to\infty$, learning rate $\eta_t\propto t^{-1}$, standard deviation $\sigma_t\propto t^{-1/2}$, loss $\overline{\ell_{f,t}}\propto t^{-1}$, $I_t=1/\sigma_t^2\propto t$ ($I_t$ is the fisher-information of the Gaussian mean).

{\bf Remark 2: non-continuity at $t=0$}. An interesting observation is that $\eta_0\neq \eta$ but rather $\eta_0\approx \frac{\eta}{2}$. This makes sense because neither $\eta_0=\eta$ nor $\eta_0=0$ leads to a loss decrease. The non-continuity goes against the common wisdom of continuous learning rate schedules.

{\bf Remark 3: decay time is bounded when $\eta\to \infty$}. Suppose we want to decrease the learning rate from the stable value $\eta$ to the final value $\eta_{\rm min}$. The optimal decay takes time $T_d = \frac{2}{a\eta_{\rm min}}(1-\frac{\eta_{\rm min}}{\eta})<\frac{2}{a\eta_{\rm min}}$. Notice that the decay time $T_d$ has an upper bound $T^*\equiv \frac{2}{a\eta_{\rm min}}$ independent of $\eta$, meaning that as long as one uses enough time (more than $T^*$) for decay, one can in principle arbitrarily increase the stable learning rate $\eta$ without worrying about extra losses induced by insufficient decay.

{\bf Remark 4: The optimal schedule for SignGD is the same except for $t_h$}. The optimal $\eta$ decay schedule for SignGD has the same functional form $\eta_t =\frac{\frac{\eta}{2}}{1+\frac{t}{t_h}}$ although with a slightly different $t_h\equiv \sqrt{2\pi}\frac{\sigma_g}{a\eta}$. Detailed derivations are deferred to Appendix~\ref{app:signgd}.

\subsection{Thermodynamics: Fourier's conduction law, Second law of thermodynamics}\label{subsec:fourier-law}

In the analysis above, the learning rate $\eta$ has two roles: temperature and time scale, making it complicated to make a correspondence with thermodynamics. We now study a simplified two-temperature setting where the analogy becomes clearer. Suppose the fast parameter $x$ reaches its thermal equilibrium with learning rate $\eta_A$. At $t=0$, we suddenly switch the learning rate to $\eta_B<\eta_A$. How does thermal width $\sigma_t$ and thermal loss $\overline{\ell_{f,t}} \equiv \frac{1}{2}a\sigma_t^2$ evolve in time?

Recalling basic facts for SGD at steady distribution: in the flat limit $a\ll \frac{2}{\eta}$, we have $\sigma \approx \sqrt{\frac{\eta}{2a}}\sigma_g$, and $\overline{\ell_f}=\frac{1}{2}a\sigma^2=\frac{1}{4}\eta\sigma_g^2$. At $t=0$, we have $\sigma_0 = \sqrt{\frac{\eta_A}{2a}}\sigma_g$. $\{\sigma_t\}$ evolves as follows $\sigma_{t+1}^2 =  (\cancel{a^2\sigma_t^2}+\sigma_g^2)\eta_B^2 - 2a\sigma_t^2\eta_B + \sigma_t^2$, 
where the crossed-out term can be ignored due to the flat limit. 
We are interested in how $\overline{\ell_{f,t}}$ evolves:
\begin{equation}\label{eq:fourier_conduction_analogy}
    \overline{\ell_{f,t+1}}-\overline{\ell_{f,t}} = \frac{1}{2}a(\sigma_{t+1}^2-\sigma_t^2)=\frac{1}{2}a\eta_B(\sigma_g^2\eta_B-2a\sigma_t^2)=-2a\eta_B( \overline{\ell_{f,t}}-\overline{\ell_{eq}}(\eta_B))
\end{equation}
where $\overline{\ell_{eq}}(\eta_B)\equiv \frac{\eta_B\sigma_g^2}{4}$ is the averaged thermal loss given $\eta=\eta_B$ in equilibrium. This equation is similar to {\bf Fourier's law} in thermal conduction $Q=k(T_A-T_B)$: when a hot object (temperature $T_A$) touches a cooler surface (temperature $T_B$), the power of thermal conduction $Q$ is proportional to their temperature difference, leading to an exponential convergence. Similarly, Eq.~(\ref{eq:fourier_conduction_analogy}) bears an exponential decay solution
$\overline{\ell_{f,t}} = \overline{\ell_f}(\eta_B) + (\overline{\ell_f}(\eta_A)-\overline{\ell_f}(\eta_B)){\rm exp}(-2a\eta_B t)\geq \overline{\ell_f}(\eta_B).$
The inequality $\overline{\ell_{f,t}}\geq \overline{\ell_f}(\eta_B)$ is related to the {\bf second law of thermodynamics}. Simply put, when a hot object is in contact with a cool thermostat, the hot object cannot be cooled down to a temperature lower than the temperature of the cool thermostat (without extra work). As a sanity check, we show in Appendix~\ref{app:fourier} that: by making Eq.~(\ref{eq:fourier_conduction_analogy}) continuous, we can also obtain the $1/t$ schedule as in Eq.~(\ref{eq:optimal_schedule}).

\subsection{Experiments}

\begin{figure}[h] 
    \centering
    \subfigure[$\ell(t_h,b)$]{
        \includegraphics[width=0.24\textwidth]
        {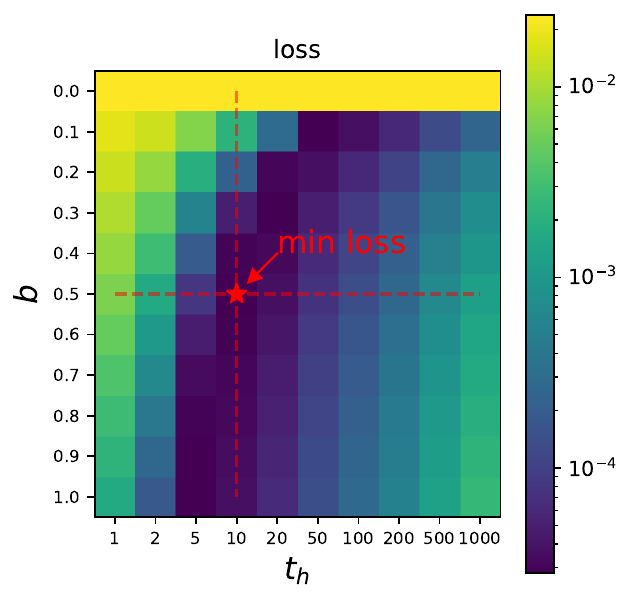}}
    \hfill
    \subfigure[$\ell(t_h=10,b)$]{
        \includegraphics[width=0.24\textwidth]
        {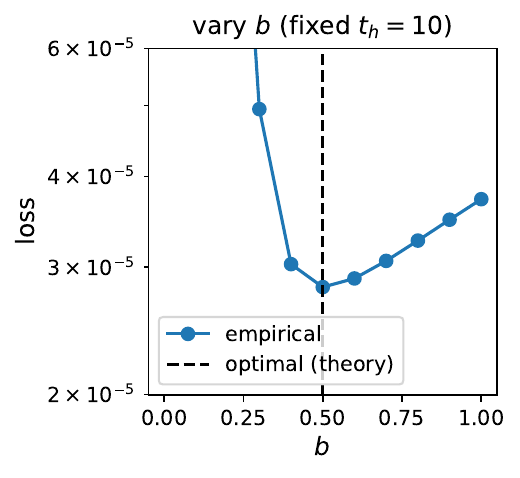}}
    \hfill
    \subfigure[$\ell(t_h,b=0.5)$]{
        \includegraphics[width=0.22\textwidth]
        {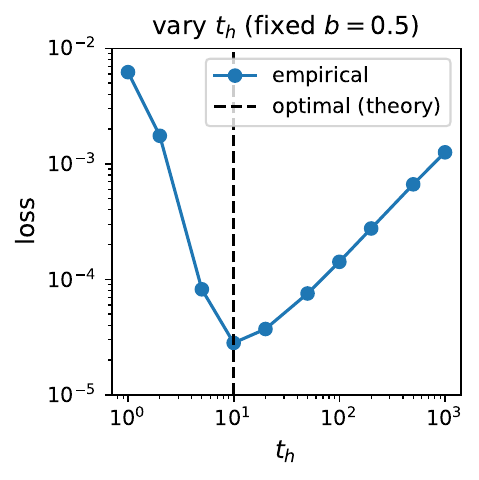}}
    \hfill
    \subfigure[$\ell(a)$]{
        \includegraphics[width=0.24\textwidth]
        {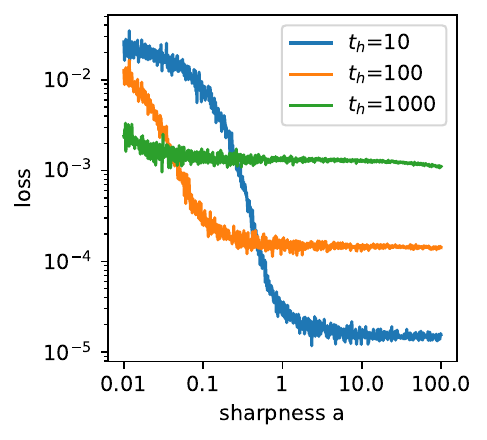}}
    \caption{Annealing toy examples. (a)(b)(c) Isotropic loss $\ell = \sum_{i=1}^n \frac{1}{2}a\theta_i^2\ (a=2, n=10000)$. The final loss obtained by applying the decay schedule $\eta_t=b\eta_0/(1+t/t_h)$. The theoretical minimum $(b,t_h)=(0.5,10)$ (marked as a star) agrees with numerical results. (d) Anisotropic loss $\ell=\sum_{i=1}^n\frac{1}{2}a_i\theta_i^2\ (a_i=10^{-2+4i/n}, n=10000)$. We set $b=0.5$ and try $t_h=10,100,1000$. Small sharpness is slower to converge than large sharpness.}
    \label{fig:optimal_decay}
\end{figure}

{\bf Toy experiments} We test the optimality of the schedule we derived in Eq.~(\ref{eq:optimal_schedule}). We choose the loss landscape $\ell = \sum_{i=1}^n \frac{1}{2}a\theta_i^2\ (a=2, n=10000)$ and initialize $\theta_i\sim \mathcal{N}(0,1)$. We run SGD with $\eta=0.1$ (manually injecting Gaussian gradient noise $\sigma_g=0.1$) for 10000 steps to reach its steady distribution. We then perform a learning rate decay schedule $\eta_t = \frac{b\eta}{1+\frac{t}{t_h}}$ with
various combinations of $(b, t_h)$. We plot the final losses in Figure~\ref{fig:optimal_decay} (a). Our theoretical result implies that $b=1/2$ and $t_h=\frac{2}{a\eta}=10$ give the best result (the lowest loss), which is verified by the phase diagram. We also plot the two slices along $t_h$ and $b$ in (b) and (c), showing that the final loss is more sensitive to $t_h$ than $b$.
To simulate anisotropy, we also experiment with an anisotropic loss $\ell=\sum_{i=1}^n\frac{1}{2}a_i\theta_i^2\ (a_i=10^{-2+4i/n}, n=10000)$, $\theta_i\sim\mathcal{N}(0,1)$. We apply the learning rate decay schedule $\eta_t = \frac{b\eta}{1+\frac{t}{t_h}}$ with $b=0.5$ and $t_h=10,100,1000$, as shown in Figure~\ref{fig:optimal_decay} (d). The loss is roughly constant for large $a$, because of the equipartition theorem in Section~\ref{sec:3}. However, small sharpness directions have higher losses because their optimal decay time $t_h\propto 1/a$ is larger, hence requiring longer time to converge.

{\bf Implications for LLM} The above results can provide insights on recent observations: (1) it is observed that the 1-sqrt decay~\cite{hagele2024scaling} or an optimized decay~\cite{luo2025a} are better than the linear or cosine decay. These better decay schedules are aligned with our derived $1/t$ schedule. (2) Decaying to zero is sub-optimal, as observed in~\cite{modded-nanogpt}. In fact, the optimal schedule implies that it should take infinite time to reach $\eta_{\rm min}=0$. 

\section{River Dynamics}\label{sec:5}

So far, we have been studying the fast dynamics of $x$, assuming the slow variable $y$ is fixed. This section will study how the slow dynamics can be influenced by the fast dynamics via entropic forces. 

\subsection{Toy model: entropic forces}

Recall that our 2D toy river-valley landscape is $l(x,y) = \frac{1}{2}a(y)x^2 + c(y)\equiv \ell_f+\ell_s$ where $\ell_f$ and $\ell_s$ are fast loss and slow loss, respectively.
The sharpness of the valley is controlled by $a(y)$, while $c(y)$ controls the bottom of the valley. Optimizer dynamics can be viewed as fast dynamics along $x$ and slow dynamics along $y$. Given a fixed $y$, the steady distribution for $x$ is $p_y(x)=\frac{1}{\sqrt{2\pi} \sigma(y)}e^{-\frac{x^2}{2\sigma(y)^2}}$. We have shown that $\sigma(y)=d(\eta, \sigma_g)/\sqrt{a(y)}$, where $d=\sqrt{\eta/2}\sigma_g$ for SGD and $c=(\pi/8)^{1/4}\sqrt{\sigma_g\eta}$ for SignGD. The entropic force is defined as the average gradient of $\ell_f$ along $y$: 
\begin{equation}
    F_{\rm ent} = -\overline{g_y} = -\frac{1}{2}a'(y)\overline{x^2} = -\frac{1}{2}a'(y)\sigma(y)^2=-\frac{d^2(\eta_,\sigma_g)}{2}\frac{a'(y)}{a(y)}
\end{equation}
The minus sign means that the entropic force points towards the direction of \textit{decreasing} sharpness. The negative gradient of the valley bottom $\ell_s$ is $F_{\rm btm} = - c'(y)$. The total ``force" is 
$F=F_{\rm ent}+F_{\rm btm}$.

{\bf Defining entropy} Notice that $a'(y)/a(y)$ in $F_{\rm ent}$ can be written into a more compact form $({\rm log}\ a(y))'$. We can define $S \equiv -\frac{d^2(\eta,\sigma_g)}{2}{\rm log}\ a(x)$, and then $F_{\rm ent}=\nabla S$. 

{\bf Entropic trapping} is defined as $F_{\rm btm}\cdot F\leq 0$. Entropic trapping happens when the entropic forces prevent the optimizer from descending the river despite the fact that the optimizer is able to ``see'' the correct direction. Concrete examples are discussed in Appendix~\ref{app:entropic-trapping}.

\subsection{Thermodynamics: entropy, the third law of thermodynamics}
We want to justify a bit more from the thermodynamics perspective why $S(x)\propto  -\frac{1}{2}{\rm log}\ a(x)$ can be interpreted as entropy. In physics, entropy is defined as $S_{\rm phy}=-\sum_i p_i{\rm log}\ p_i$ for discrete systems or $S_{\rm phy}=-\int dx\ p(x){\rm log}\ p(x)$ for continuous systems. A Gaussian distribution $p(x)=\frac{1}{\sqrt{2\pi} \sigma}e^{-\frac{x^2}{2\sigma^2}}$ hence has entropy $S_{\rm phy}=\frac{1}{2}{\rm log}(2\pi\sigma^2)+\frac{1}{2}$. Inserting $\sigma=d/\sqrt{a}$ gives $S_{\rm phy} = -\frac{1}{2}{\rm log}\ a+ \frac{1}{2}{\rm log}(2\pi d)+\frac{1}{2}$, which scales with $a$ as $S\propto -\frac{1}{2}{\rm log}\ a$. This is also analogous to the {\bf Boltzmann entropy equation} $S=k_b{\rm log} W$, where $k_b$ is the Boltzmann constant, and $W$ is the number of microstates (analogous to $\sigma$). The Boltzmann entropy equation is a foundation to {\bf the third law of thermodynamics}.

\subsection{Experiments}

{\bf LLM experiments} We want to determine to what extent the concerning entropic trapping phenomenon occurs in LLM training. Technically, the entropic force is hard to compute for large models: by definition, entropic forces are third-order derivatives, since they are gradients of sharpness (second-order derivatives). However, we may probe the existence of entropic forces via loss curves alignment against learning rate sums. In the gradient flow limit $(\eta\to 0)$, the slow dynamics is only governed by the learning rate sum~\cite{luo2025a}, i.e., we should expect the following two schedules to give the same final state and loss:
(i) learning rate $\eta$ with $A$ steps and (ii) learning rate $2\eta$ with $A/2$ steps. However, for a finite $\eta$, the two schedules may not align due to the existence of entropic forces, so their misalignment is an indicator of the magnitude and direction of entropic forces. Since $\ell_s$ is not directly measurable, we can try to control $\ell_f$ the same (requiring $\eta_{\rm min}$ to be the same, according to Section~\ref{sec:4}) and measure $\ell$ as a surrogate of $\ell_s$.


\begin{figure}
    \centering
    \includegraphics[width=0.8\linewidth]{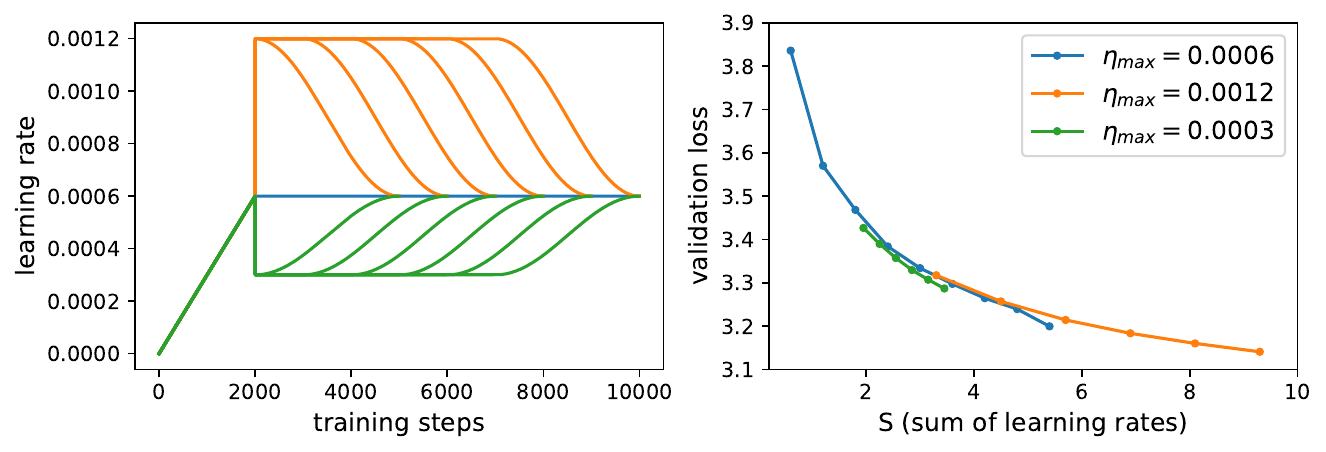}
    \caption{Test the existence of entropic forces in LLMs. Left: Various learning rate schedules with different stable $\eta_{\rm max}=0.0003,0.0006,0.0012$ and the same $\eta_{\rm min}=0.0006$. Right: Plot validation losses against learning rate sums. Curves for different $\eta$ roughly align, suggesting slightly negative entropic forces, corresponding to a slightly narrowing valley along the river.}
    \label{fig:loss_S_etamax}
    \vskip -0.2cm
\end{figure}

We test a bunch of schedules with different stable learning rates $\eta=0.0003, 0.0006,0.0012$ (Figure~\ref{fig:loss_S_etamax} left). The stable phase may last for $1000i$ steps $(i=0,1,2,3,4,5)$, and the decay phase lasts for 3k steps with cosine transition to the final learning rate $\eta_{\rm min}=0.0006$\footnote{Although $\eta_{\rm min}$ has the subscript ``min'', it does not necessarily mean it is the minimum learning rate in the last phase. The last phase can either be decay ($\eta=0.0012$), stable ($\eta=0.0006$) or growth ($\eta=0.0003$). The point is that all the schedules have the same final learning rate, denoted as $\eta_{\rm min}$.}. For each $\eta$, we plot its validation loss against the $\eta$ sum, swept by varying the duration of the stable phase. Figure~\ref{fig:loss_S_etamax} (b) shows that curves of different $\eta$ align reasonably well with each other, although smaller $\eta$ seem to produce slightly lower losses given the same $\eta$ sum, similar to the observations in~\cite{luo2025a}. This result implies that LLM has a slightly narrowing valley structure on average (correspondingly, the entropic force is slightly negative). 
However, our training is in the early stage due to computational constraints. It would be interesting to study whether entropic forces represent a more significant limit to the training of LLMs when more training steps are taken (where valleys may become sharper and entropic forces are larger).

\section{Summary of Findings}\label{sec:6}

{\bf The role of learning rates} We have learned that the learning rate $\eta$ has three roles in controlling training dynamics. (1) {\bf Gaussian width}: $\eta$ acts as temperature, controlling the Gaussian width along the fast direction. (2) {\bf Entropic force magnitude}: the Gaussian width, combined with valley sharpness, jointly controls the entropic force. (3) {\bf Time scale}: controls the step size.

{\bf What determines the final loss?} In summary, our results show that the final loss largely depends on learning rate sum $D$ (controlling $\ell_s$, Section~\ref{sec:5}) and $\eta_{\rm min}$ (controlling $\ell_f$, Section~\ref{sec:3}), but there are also small correction terms due to the entropic force $\Delta_{\rm entropic}$ (Section~\ref{sec:5}) and due to insufficient annealing $\Delta_{\rm anneal}$ (Section~\ref{sec:4}).
\begin{equation}
    \ell_{\rm final}=\ell(D,\eta_{\rm min}) + \Delta_{\rm entropic} + \Delta_{\rm anneal}.
\end{equation}
Empirically, for the early training stages of GPT2, we found $\Delta_{\rm entropic}\sim 0$ holds roughly true, and $\Delta_{\rm anneal}\sim 0$ when the decay phase is no smaller than 3k steps. If we assume $\Delta_{\rm entropic}$ and $\Delta_{\rm anneal}$ can be ignored, the only way to reduce loss is by reducing $\ell(D,\eta_{\rm min})$, which involves reducing $\eta_{\rm min}$ and/or increasing $D$. Increasing $D$ can be achieved by using a larger step size in the stable phase, verified by experiments in Appendix~\ref{app:lr_schedule}.

\section{Related Works}\label{sec:related_works}

{\bf Physics of optimization} Stochastic gradient descent with finite step sizes has different dynamics from gradient flow. Finite learning rate $\eta$ can induce implicit regularization $\frac{\eta}{4}||\nabla \ell||^2$~\cite{barrett2020implicit}, and stochastic gradients also have various implicit biases~\cite{kunin2023limiting,chen2023stochastic,stephan2017stochastic}, guiding the optimization towards flatter minima~\cite{xie2020diffusion,cohen2021gradient}, large eigendirections~\cite{,wu2020direction},  maximizes margin~\cite{lyu2019gradient,wang2021implicit}, and redundant neurons/directions~\cite{chen2023stochastic,xu2025overview}. Besides these quite general phenomena, research has also been carried out to understand the loss landscapes of neural networks, especially in the over-parametrized regime. Large models are shown to have mode connectivity~\cite{garipov2018loss,frankle2020linear}, which is also related to the recently discovered river-valley landscape of LLMs~\cite{wen2024understanding,liu2025focus}.

{\bf Learning rate schedules} for LLMs are diverse: cosine decay~\cite{loshchilov2016sgdr}, cyclic~\cite{smith2017cyclical}, Noam~\cite{vaswani2017attention} and weight-stable-decay (WSD)~\cite{hu2024minicpm}. Recent research has started to show the advantages of the WSD schedule~\cite{wen2024understanding, luo2025a, hagele2024scaling} and concerns about designing better decay schedules, e.g., the 1-sqrt schedule proposed in~\cite{hagele2024scaling} and an optimized schedule in~\cite{luo2025a}. Our analysis provides yet another theoretical evidence for the use of the WSD schedule and analytically derives an optimal decay schedule which decays as $1/t$ (under the isotropic assumption).

{\bf Thermodynamics and learning} Although we are the first to establish a mapping between thermodynamics and LLM training dynamics, thermodynamics has long inspired and has connections to machine learning: optimization as a thermodynamic process~\cite{welling2011bayesian}, statistical mechanics of learning~\cite{bahri2020statistical}, information bottleneck~\cite{tishby2000information}, entropy gradient descent~\cite{chaudhari2019entropy}, Boltzman machine~\cite{ackley1985learning}, Hopfield networks~\cite{ramsauer2020hopfield}, diffusion models~\cite{sohl2015deep,song2020score,ho2020denoising}, thermodynamic interpretations of networks~\cite{mehdi2024thermodynamics}.

\section{Conclusions}\label{sec:conclusions}

We propose a toy model of a river–valley loss landscape and analyze the training dynamics under SGD and SignGD. The fast–slow separation enables us to treat valley and river directions independently, yielding analytically tractable results: thermal equilibrium and annealing for the fast dynamics, and drift for the slow dynamics. These analytical solutions bear qualitative—and in some cases quantitative—analogies to classical thermodynamic concepts and laws. Crucially, they are relevant to large language model (LLM) training, as recent work has shown that LLM loss landscapes exhibit river–valley structure. This duality between optimization and thermodynamics offers a novel perspective for understanding and evaluating modern optimizers. While we leave it for future work, we include a proof-of-concept analysis in Appendix~\ref{app:focus}, where we analyze the recently proposed FOCUS optimizer—characterized by self-attracting forces~\cite{liu2025focus}—through the lens of our theory.

{\bf Limitations}. Many of the derivations in this paper adopt the physicist's style of reasoning—emphasizing intuition, simplification, and tractable approximations—which may not satisfy the standards of mathematical rigor expected by theorists. For instance, the Gaussian approximation of steady states is not necessary for many results (only variances matter). In deriving the optimal learning rate decay schedule, we assume either uniform sharpness or a one-dimensional landscape; we treat the river as straight, though it is likely curved in practice; and we ignore momentum and weight decay for simplicity. Despite these simplifications, our analysis yields non-trivial, testable insights into LLM training dynamics. Natural extensions of this work include relaxing these assumptions, validating predictions at larger scales, and generalizing the framework. Although our focus is on transformer-based LLMs, the underlying physics-inspired principles may extend to other model architectures.

{\bf Acknowledgment}  Z.L. and M.T. are supported by IAIFI through NSF grant PHY-2019786. Z. L. is also supported by the Google PhD fellowship.

\bibliographystyle{unsrt}
\bibliography{ref.bib}

\newpage
\appendix

{\huge Appendix}

\section{SGD converges to Gaussian steady distribution}\label{app:sgd-converge-gaussian}

Suppose the initial point is $x_0$ at $t=0$. The distribution is a delta function, effectively a Gaussian with mean $\mu_0=x_0$ and $\sigma_0=0$. Due to the linearity of Eq.~(\ref{eq:sgd}), a Gaussian distribution at step $t$ evolved to become a Gaussian distribution at step $t+1$, although with different means $\mu_t$ and standard deviations $\sigma_t$. Their recursive relations are ($t=1,2,3,\cdots$) 
\begin{equation}
    \begin{aligned}
    &\mu_{t} = (1-\eta a)\mu_{t-1}, \\
    &\sigma_t^2=(1-\eta a)^2\sigma_{t-1}^2+\eta^2\sigma_g^2,
\end{aligned}
\end{equation}
which have solutions
\begin{equation}
    \begin{aligned}
    &\mu_t = (1-\eta a)^{t} x_0, \\
    &\sigma_t^2 = (1-(1-\eta a)^{2t})\sigma^2, \quad \sigma\equiv\frac{\sigma_g}{\sqrt{a(\frac{2}{\eta}-a)}},
    \end{aligned}
\end{equation}
where $\sigma$ is the steady-state standard deviation. Regardless of $x_0$, $\mu_t\to 0$ and $\sigma_t\to \sigma$ as $t\to\infty$. The time scale of convergence is $t_c=-1/{\rm log}(1-\eta a)$. In the flat limit $a\eta\ll 1$, $t_c\approx 1/(a\eta)$.

\section{Derivations for SGD}

\subsection{Fixed learning rate}

Suppose a 1D loss function $l(x)=\frac{1}{2}ax^2$ where $a$ is the second order derivative of the quadratic function. An SGD optimizer with learning rate $\eta$ and gradient noise $\sigma_g$ obeys the following dynamics:
\begin{equation}\label{eq:sgd_app}
    x_{t+1} = x_t - \eta(a x_t + \sigma_g\dot{W}_t)
\end{equation}
The equilibrium distribution is the Gaussian distribution $p(x)=\frac{1}{\sqrt{2\pi}\sigma}e^{-\frac{x^2}{2\sigma^2}}$. In equilibrium, we should have ${\rm Var}(x_{t+1})={\rm Var}(x_t)$, i.e.,
\begin{equation}
    \sigma^2 = (1-\eta a)^2\sigma^2 + \sigma_g^2,
\end{equation}
which gives
\begin{equation}\label{eq:sgd_sigma}
    \sigma = \frac{\sigma_g}{\sqrt{a(\frac{2}{\eta}-a)}}.
\end{equation}
This formula is only valid when $0<a<\frac{2}{\eta}$. When $a\to 0$ (i.e., flat), finite $\sigma_g$ can induce infinite $\sigma$ (i.e., steady distribution does not exist). When $a\to \frac{2}{\eta}$, learning rate reaches the so-called ``edge of stability''. In particular, when $\eta>\frac{2}{a}$, the $\{x_t\}$ sequence diverges when $\sigma_g\to 0$. 

\begin{figure}[ht]
    \centering
    \includegraphics[width=0.45\linewidth]{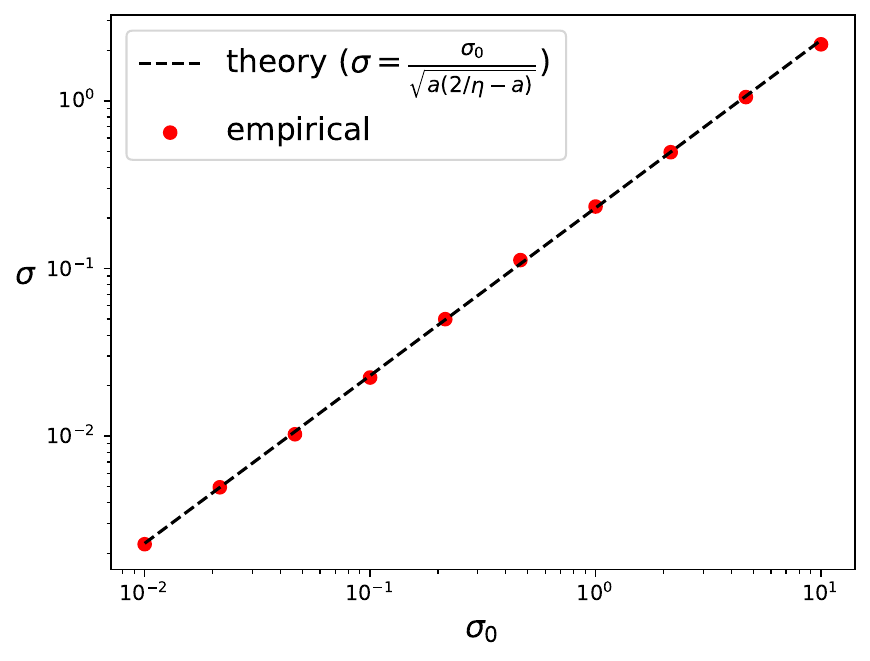}
    \includegraphics[width=0.45\linewidth]{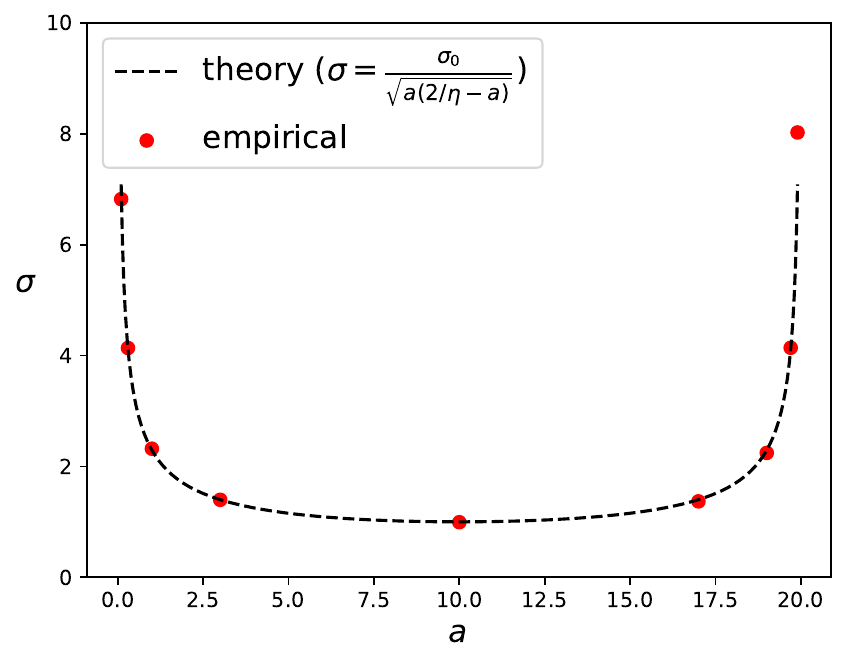}
    \caption{Dependence of $\sigma$ on gradient noise $\sigma_g$ and sharpness $a$.}
    \label{fig:sigma_dependence}
\end{figure}

Eq.~\ref{eq:sgd_sigma} is tested empirically in Figure~\ref{fig:sigma_dependence}.
A somewhat unexpected feature of Eq.~(\ref{eq:sgd_sigma}) is that $\sigma\to \infty $  when $a\to 0$ although we were previously viewing flat directions as good and viewing sharp directions as evil. Eq.~(\ref{eq:sgd_sigma}) suggests that flat directions are as evil as sharp directions. 

{\bf Equipartition theorem} The original idea of equipartition (in classical statistical mechanics) is that, in thermal equilibrium, energy is shared equally among all of its degrees of freedom. Specifically, each degree of freedom would contribute to energy $\frac{1}{2}k_B T$ ($k_B$: Boltzmann constant, $T$: temperature) regardless of underlying details. We show that the loss incurred due to gradient noise is also (approximately) independent of sharpness $a$: $\langle l\rangle = \frac{1}{2}a\langle x^2\rangle=\frac{1}{2}a\sigma^2$. When $a\eta \ll 2$, $\sigma\approx \frac{1}{\sqrt{2a}}\sigma_g\sqrt{\eta}$. So $\langle l\rangle=\frac{1}{2}a\sigma^2\approx \frac{1}{2}a(\frac{1}{\sqrt{2a}}\sigma_g\sqrt{\eta})^2=\frac{1}{4}\sigma_g^2\eta$. Ignoring constants, the effective temperature is $T_{\rm eff}\propto \sigma_g^2\eta$. The gradient noise scales with batch size $B$ as $\sigma_g\propto \frac{1}{\sqrt{B}}$. To reduce temperature, we can increase the batch size or decrease the learning rate. This has an interesting implication: during the training of a neural network, there might exist many such equilibrium directions. No matter how sharpness these directions are, they contribute equally to the total loss. Suppose there are $N$ such directions, the total loss incurred by gradient noise is $l_g=N\langle l\rangle=\frac{1}{4}N\sigma_g^2\eta$.

\subsection{Learning rate decay}
Now we consider a learning rate schedule, i.e., a sequence of $\{\eta_t\}_{t=0}^T$. Eq.~(\ref{eq:sgd}) now becomes:
\begin{equation}
    x_{t+1} = x_t - \eta_t(a x_t + \sigma_g\dot{W}_t)
\end{equation}
Since the equation is linear, if $p(x_0)$ starts off as a Gaussian distribution, $p(x_t)$ remain a Gaussian distribution (with time-varying Gaussian width, denoted as $\sigma_t$) forever.
Assuming that at $t=0$, $p(x_0)$ is already in thermal equilibrium whose Gaussian width is given by Eq.~(\ref{eq:sgd_sigma}), i.e., the initial condition is $\sigma_0=\sigma$. Gaussian widths $\sigma_t$ obeys the following recursive relation:
\begin{equation}
    \sigma_{t+1}^2 = (1-\eta_t a)^2\sigma_t^2 + (\eta_t\sigma_g)^2 = (a^2\sigma_t^2+\sigma_g^2)\eta_t^2 - 2a\sigma_t^2\eta_t + \sigma_t^2,
\end{equation}
which is a quadratic function of $\eta_t$. We want to choose $\eta_t$ such that $\sigma_{t+1}$ is minimized:
\begin{equation}
    \eta_t = \frac{a\sigma_t^2}{a^2\sigma_t^2+\sigma_g^2},\quad  \sigma_{t+1}^2 = \frac{\sigma_g^2\sigma_t^2}{\sigma_g^2+a^2\sigma_t^2}.
\end{equation}
By inverting the second equation, we get
\begin{equation}
    \frac{1}{\sigma_{t+1}^2} = \frac{1}{\sigma_t^2} + \frac{a^2}{\sigma_g^2},
\end{equation}
which means that $\{\frac{1}{\sigma_t^2}\}$ forms an arithmetic sequence. It is clear that $\frac{1}{\sigma_t^2}=\frac{1}{\sigma_0^2} + \frac{a^2t}{\sigma_g^2}$. Correspondingly,
\begin{equation}
    \eta_t = \frac{\frac{1}{a}}{1+\frac{\sigma_g^2}{a^2\sigma_t^2}} = \frac{\frac{1}{a}}{1+\frac{\sigma_g^2}{a^2}(\frac{1}{\sigma_0^2}+\frac{a^2t}{\sigma_g^2})} \approx \frac{\frac{1}{a}}{\frac{2}{a\eta}+t}\propto \frac{1}{t+t_h},\quad (t_h\equiv \frac{2}{a\eta})
\end{equation}
whose asymptotic behavior is $\eta_t\propto \frac{1}{t}$. $t_h$ is the time needed to decrease $\eta_t$ by half, hence representing the characteristic time scale of learning rate decay. Another interesting observation is that $\eta_0\neq \eta$. In fact, when we assume $a\eta\ll 2$, $\eta_0\approx \frac{\eta}{2}$. This makes sense: since $\sigma_1$ is an quadratic function of $\eta_0$, and $\sigma_1=\sigma_0$ for both $\eta_t = \eta$ (continue thermal equilibrium) and $\eta_0=0$ (freeze), the best $\eta_1$ must be at the middle point of $\eta$ and $0$, which is $\frac{\eta}{2}$. This suggests that the standard learning rate decay schedule (which is continuous at $t=0$) may be suboptimal. 

$\eta_t = \frac{\eta t_h}{t+t_h} (t_h=\frac{2}{a\eta})$, to take it from $\eta$ to $\eta_m$, it takes time $T_d = \frac{2}{a\eta_m}(1-\frac{\eta_m}{\eta})<\frac{2}{a\eta_m}$. So a larger $\eta$ in the stable phase does not increase $T_d$ significantly.

\section{Derivations for SignGD}\label{app:signgd}

\subsection{Fixed learning rate}

\begin{figure}[ht]
    \centering
    \includegraphics[width=0.45\linewidth]{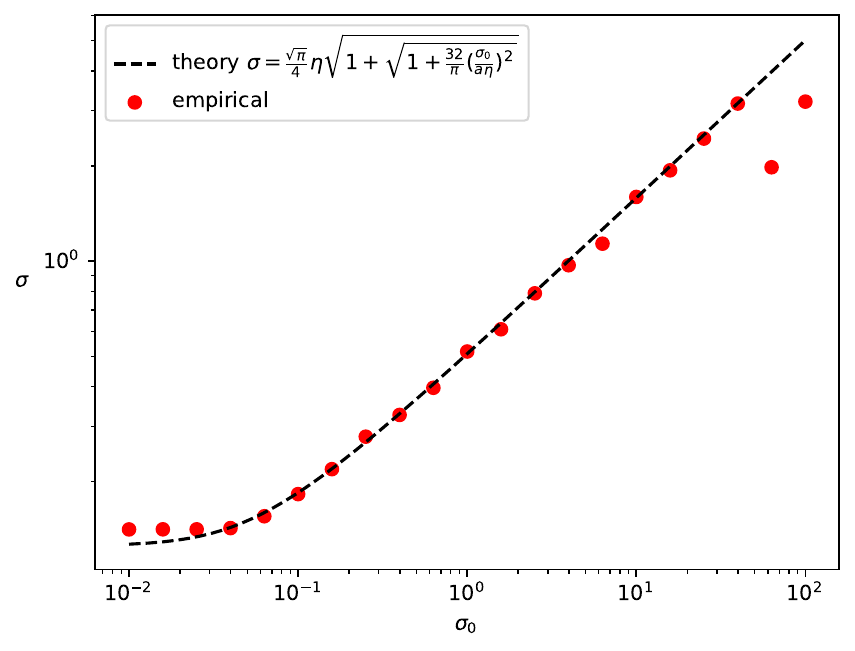}
    \includegraphics[width=0.45\linewidth]{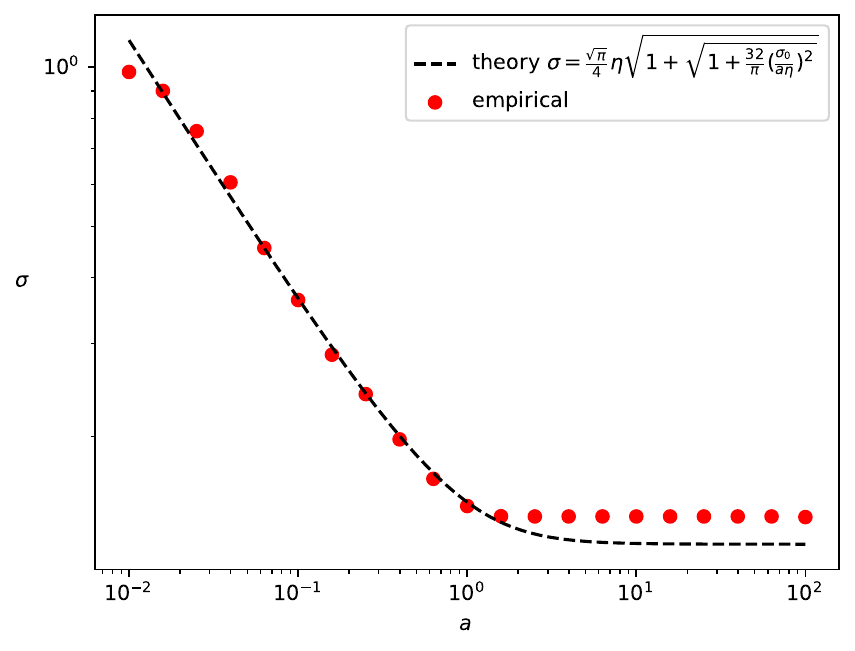}
    \caption{SignGD: Dependence of $\sigma$ on gradient noise $\sigma_g$ and sharpness $a$.}
    \label{fig:signgd_sigma_dependence}
\end{figure}

The optimization dynamics of SignGD is 
\begin{equation}\label{eq:signgd_app}
    x_{t+1} = x_t - \eta\ {\rm sign}(ax_t+\sigma_g\dot{W}).
\end{equation}
Define $\zeta(x)\equiv \int_{-ax/\sigma_g} \frac{1}{\sqrt{2\pi}}{\rm exp}(-\frac{y^2}{2})dy$. Then given a fixed $x$, $ax+\sigma_g\dot{W}$ is positive with probability $\zeta(x)$, or negative with probability $1-\zeta(x)$, i.e.,
\begin{equation}
    x_{t+1}=
\begin{cases}
  x_t-\eta & {\rm probaility}\ \zeta(x) \\
  x_t+\eta & {\rm probaility}\ 1-\zeta(x)\\
\end{cases}
\end{equation}
We assume that $x_t$ has a Gaussian steady distribution $x_t\sim \frac{1}{\sqrt{2\pi}\sigma}{\rm exp}(-\frac{x^2}{2\sigma^2})$~\footnote{The Gaussian approximation becomes increasingly more accurate as $\eta\to 0$.}. We can set up an equation by leveraging ${\rm Var}(x_{t+1})={\rm Var}(x_t)$, i.e.,
\begin{equation}
    \int_{-\infty}^\infty dx\ p(x)x^2 = \int_{-\infty}^\infty dx\ p(x)(\zeta(x)(x-\eta)^2+(1-\zeta(x))(x+\eta)^2),
\end{equation}
which can simplify to (with some derivations)
\begin{equation}
    \int_{-\infty}^\infty dx\ p(x)\zeta(x)x - \frac{\eta}{4}=0.
\end{equation}
The first integral 
\begin{equation}
    \begin{aligned}
        \int_{-\infty}^\infty dx\ p(x)\zeta(x)x = & \int_{-\infty}^\infty dx\ \int_{-ax/\sigma_g}^\infty dy\ \frac{x}{\sqrt{2\pi}\sigma}{\rm exp}(-\frac{x^2}{2\sigma^2})\frac{1}{\sqrt{2\pi}}{\rm exp}(-\frac{y^2}{2})\\
         = & \int_{-\infty}^\infty\ dy\int_{-\sigma_gy/a}^\infty\ dx\frac{x}{\sqrt{2\pi}\sigma}{\rm exp}(-\frac{x^2}{2\sigma^2})\frac{1}{\sqrt{2\pi}}{\rm exp}(-\frac{y^2}{2}) \\
         = & \int_{-\infty}^\infty\ dy\ {\rm exp}(-\frac{y^2}{2})\int_{-\sigma_gy/a}^\infty\ dx\frac{x}{2\pi\sigma}{\rm exp}(-\frac{x^2}{2\sigma^2}) \\
         = & \int_{-\infty}^\infty\ dy\ {\rm exp}(-\frac{y^2}{2})\frac{\sigma}{2\pi}{\rm exp}(-\frac{1}{2}(\frac{\sigma_g y}{a\sigma})^2) \\
         = & \frac{\sigma}{\sqrt{2\pi}}\frac{1}{\sqrt{1+(\frac{\sigma_g}{a\sigma})^2}}
    \end{aligned}
\end{equation}
Combining the previous two equations gives
\begin{equation}
    \frac{\sigma}{\sqrt{1+(\frac{\sigma_g}{a\sigma})^2}}=\sqrt{\frac{\pi}{8}}\eta,
\end{equation}
or
\begin{equation}
    \sigma = \frac{\sqrt{\pi}}{4}\eta \sqrt{1+\sqrt{1+\frac{32}{\pi}(\frac{\sigma_g}{a\eta})^2}},
\end{equation}
which is verified in Figure~\ref{fig:signgd_sigma_dependence}. The deviation at large $a$ or small $\sigma_g$ is due to discretization ($x_t$ distribution cannot be viewed as a Gaussian). Comparing to SGD (Figure~\ref{fig:sigma_dependence}), the main difference is the $\sigma(a)$ is non-monotonic for SGD but is monotonic for SignGD. Another observation is that $\sigma$ is lower bounded by $\frac{\sqrt{2\pi}}{4}\eta$ (when $\sigma_g\to 0$ or $a\to \infty$). In the limit of $\frac{\sigma_g}{a\eta}\gg 1$, we have $\sigma\propto \frac{\sqrt{\sigma_g}\sqrt{\eta}}{\sqrt{a}}$. Comparing this with SGD:
\begin{equation}
    \sigma^{\rm SGD} \propto \frac{\sigma_g\sqrt{\eta}}{\sqrt{a}},\quad \sigma^{\rm SignGD}\propto \frac{\sqrt{\sigma_g}\sqrt{\eta}}{\sqrt{a}},
\end{equation}
which suggests that SignGD is more robust to bigger noise since $\sigma^{\rm SGD}\propto \sigma_g$ while $\sigma^{\rm SignGD}\propto \sqrt{\sigma_g}$. Also since $\sigma^{\rm SignGD}\propto\frac{1}{\sqrt{a}}$ just like $\sigma^{\rm SGD}\propto\frac{1}{\sqrt{a}}$, the equipartition property applies to SignGD as well.

\subsection{Learning rate decay}

Considering time-varying learning rate $\{\eta_t\}_{t=0}^T$, Eq.~(\ref{eq:signgd}) becomes:
\begin{equation}\label{eq:signgd_etat}
    x_{t+1} = x_t - \eta_t\ {\rm sign}(ax_t+\sigma_g\dot{W}).
\end{equation}
Similarly, we obtain the recursive relation for $\sigma_t$:
\begin{equation}
    \sigma_{t+1}^2= \eta_t^2 - 4\eta_t \int_{-\infty}^\infty dx\  p(x)\zeta(x)x + \sigma_t^2 = \eta_t^2 - \frac{4\sigma_t}{\sqrt{2\pi}\sqrt{1+(\frac{\sigma_g}{a\sigma_t})^2}}\eta_t + \sigma_t^2,
\end{equation}
which is a quadratic function against $\eta_t$. The optimal choice of $\eta_t$ that minimizes $\sigma_t$ is:
\begin{equation}
    \eta_t = \frac{2\sigma_t}{\sqrt{2\pi}\sqrt{1+(\frac{\sigma_g}{a\sigma_t})^2}},\quad \sigma_{t+1}^2=\sigma_t^2-\frac{2}{\pi}\frac{\sigma_t^2}{1+(\frac{\sigma_g}{a\sigma_t})^2}\approx \sigma_t^2 -\frac{2}{\pi}(\frac{a\sigma_t^2}{\sigma_g})^2,
\end{equation}
where the approximation holds when $\sigma_g\gg a\sigma_0\geq a\sigma_t$ which is equivalent to $\sigma_g\gg a\eta$.
Inverting both sides of the second equation gives
\begin{equation}
    \frac{1}{\sigma_{t+1}^2}=\frac{1}{\sigma_t^2} \frac{1}{1-\frac{2}{\pi}(\frac{a\sigma_t}{\sigma_g})^2}\approx \frac{1}{\sigma_t^2}(1+\frac{2}{\pi}(\frac{a\sigma_t}{\sigma_g})^2)=\frac{1}{\sigma_t^2} + \frac{2}{\pi}(\frac{a}{\sigma_g})^2,
\end{equation}
which is an arithmetic sequence. Hence $\frac{1}{\sigma_t^2} = \frac{1}{\sigma_0^2} + \frac{2}{\pi}(\frac{a}{\sigma_g})^2t$. Correspondingly
\begin{equation}
    \eta_t \approx \sqrt{\frac{2}{\pi}}\frac{a\sigma_t^2}{\sigma_g} = \frac{\sqrt{\frac{\pi}{2}}\frac{\sigma_g}{a}}{t+\sqrt{2\pi}\frac{\sigma_g}{a\eta}}\propto\frac{1}{t+t_h},\quad (t_h\equiv \sqrt{2\pi}\frac{\sigma_g}{a\eta})
\end{equation}
whose asymptotic behavior is $\eta_t\sim \frac{1}{t}$, just like SGD. We again observe that $\eta_0\approx \eta/2$, which suggests that a continuous learning rate schedule might be suboptimal. Comparing the half time of SGD and SignGD:
\begin{equation}
    t_h^{\rm SGD} = \frac{2}{a\eta},\quad t_h^{\rm SignGD} = \sqrt{2\pi}\frac{\sigma_g}{a\eta}.
\end{equation}
For a large model, the gradient (and gradient noise) of each individual parameter is usually small, i.e., $\sigma_g<1$, suggesting the benefit of SignGD over SGD.

\section{Generalizing the two-temperature setup}\label{app:fourier}
In Section~\ref{subsec:fourier-law}, we have shown that in the two-learning rate (temperature) setup $\eta_A>\eta_B$, the convergence from one equilibrium ($\eta_A$) to the other equilibrium ($\eta_B$) is exponential, similar to the situation when a hot object is in contact with a cooling thermostat, the temperature of the hot object converges exponentially to the temperature of the thermostat. However, this does not mean we can have $\overline{\ell_f}$ decay exponentially in time. What explains the difference? Note that the convergence rate $2a\eta_B$ is also proportional to $\eta_B$, meaning that $\eta_B$ play both roles: temperature and time scale. This is different from the situation in thermodynamics when temperature and time are independent (not controlled by a single parameter).

Making Eq.~(\ref{eq:fourier_conduction_analogy}) continuous, we have
\begin{equation}
    \frac{d\overline{\ell}}{dt} = -2a\eta(\overline{\ell} - \overline{\ell_{eq}}(\eta)),\quad \overline{\ell_{eq}}(\eta)\equiv C\eta.
\end{equation}
Our goal is to design $\eta(t)$ such that $\overline{\ell}(t)$ is minimized. Given $\overline{\ell}$, the RHS is an inverted quadratic function of $\eta$, so the optimal $\eta$ is $\eta^*(\overline{\ell})\equiv \overline{\ell}/(2C)$. Inserting the relation to the RHS, we have
\begin{equation}
    \frac{d\overline{\ell}}{dt} = -\frac{a}{2C}\overline{\ell}^2,
\end{equation}
which solves to be $\overline{\ell}(t)=(\overline{\ell}(0)^{-1}+\frac{at}{2C})^{-1}$, and correspondingly $\eta(t)=(\frac{2C}{\overline{\ell}_0}+at)^{-1}$.

\section{Entropic Trapping}\label{app:entropic-trapping}

\subsection{Example 1: Go down and stop}
\begin{figure}[ht]
    \centering
    \includegraphics[width=0.31\linewidth]{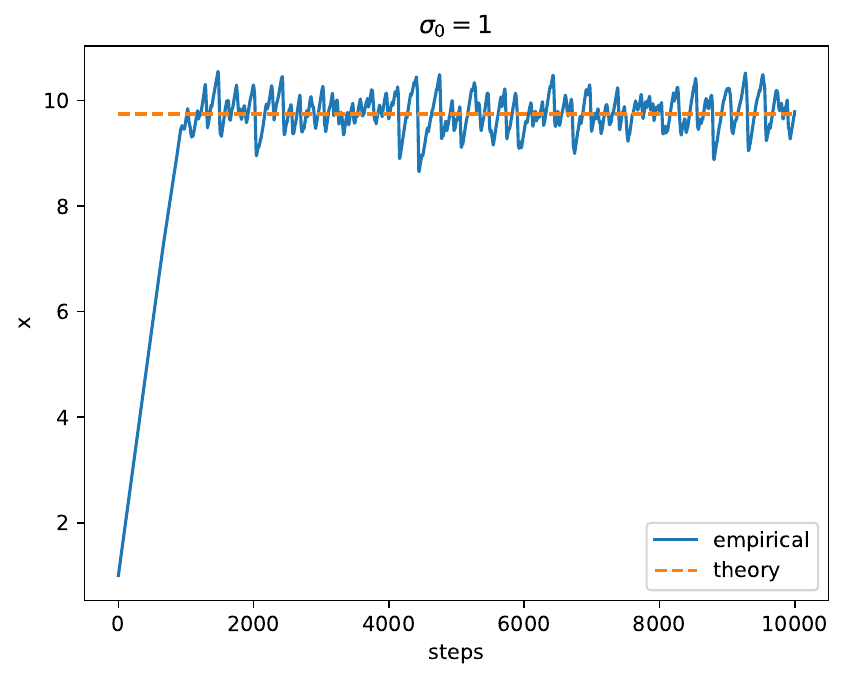}
    \includegraphics[width=0.31\linewidth]{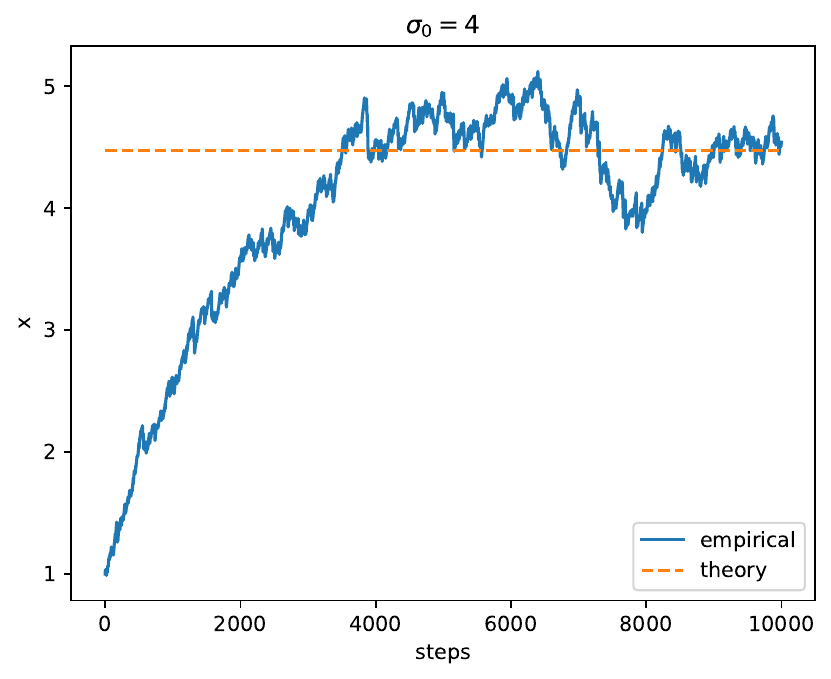}
    \includegraphics[width=0.31\linewidth]{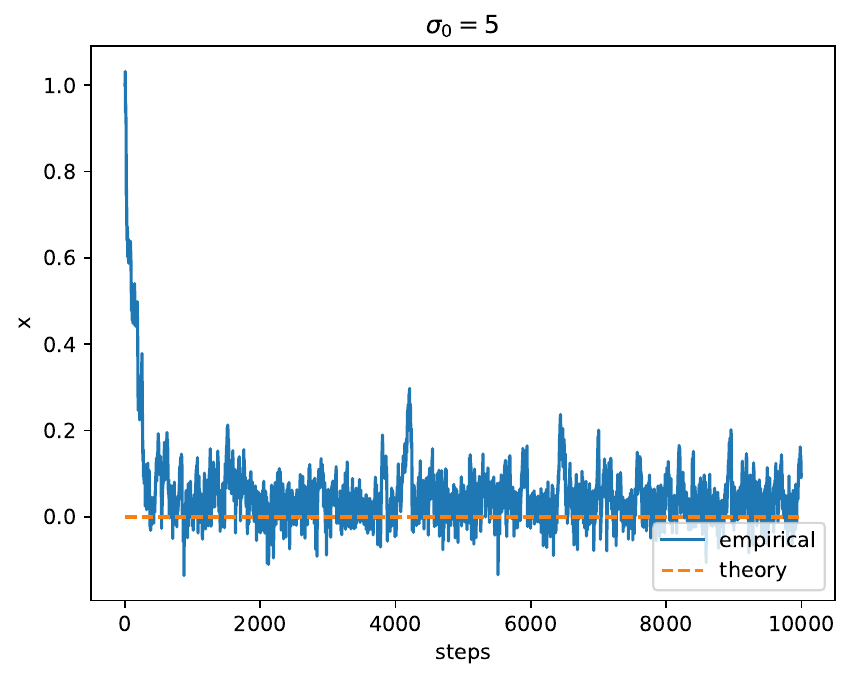}
    \caption{Increasing gradient noise $\sigma_g$ makes the stop point shift to left (flatter region).}
    \label{fig:stop_position}
\end{figure}

We set $c(x)=-cx\ (c=0.1)$, $a(x)=a_0+b|x|\ (a_0=b=1)$. Solving $F=0$ gives
\begin{equation}
    x_{-,+}=\frac{1}{\eta} \pm \sqrt{(\frac{1}{\eta})^2-\frac{b\sigma_g^2}{2c}}.
\end{equation}
When $x>x_+$ or $x<x_-$, $F(x)<0$, i.e., $x$ moves to the left. When $x_-<x<x_+$, $F(x)>0$, $x$ moves to the right. As a result, when the initial point $x_0<x_-$ or $(\frac{1}{\eta})^2<\frac{b\sigma_g^2}{2c}$, $x$ would decrease and end at 0. When $x_-<x<x_+$, $x$ would decrease and end at $x_+$. This is verified in Figure~\ref{fig:stop_position}.

\subsection{Example 2: Either left or right}

Let us consider a specific case with $a(y)={\rm exp}(ay)$ and $c(y)=-cy$. We have $F_{\rm btm}=c$ and $F_{\rm ent}=-\frac{d^2(\eta,\sigma_g)a}{2}$ which are both independent of $y$. When $d>\sqrt{\frac{2c}{a}}$, $F_{\rm ent}+F_{\rm btm}<0$, meaning that $y$ becomes smaller. $d>\sqrt{\frac{2c}{a}}$ translates to $\eta>\eta^*=\frac{4c}{a\sigma_g^2}$ (SGD) and $\eta>\eta^*\equiv\sqrt{\frac{32}{\pi}}\frac{c}{a\sigma_g}$ (SignGD). This means, to get deep down a narrowing valley, the learning rate should not be set too large, otherwise, the dynamic blocking could happen.

\section{Attraction forces}\label{app:focus}

SGD with linear attraction force:
\begin{equation}
    \bar{x}_t=\beta\bar{x}_{t-1} + (1-\beta) x_t,\ x_t = x_{t-1} - \eta(ax_t+\gamma (x_{t-1}-\bar{x}_{t})+\sigma_g\dot{W}_{t-1}),
\end{equation}
with the standard deviation of $x$ being
\begin{equation}
    \sigma = \sqrt{\frac{\eta (1-\beta^2+a\beta(1+\beta)\eta-2\beta^3\gamma\eta)}{(a+\gamma(1-2\beta))(2(1+\beta)-\eta((a+\gamma)(1+\beta)-2\beta^2\gamma))(1+\beta(-1+a\eta+2(1-\beta)\gamma\eta))}}\sigma_g.
\end{equation}
We verify that when $\gamma\to 0$, $\sigma$ goes back to $\sigma=\frac{1}{\sqrt{a(\frac{2}{\eta}-a)}}\sigma_g$.

When $\beta\to 0$, $\sigma=\frac{1}{\sqrt{(a+\gamma)(\frac{2}{\eta}-(a+\gamma))}}\sigma_g$, which means that the role of $\gamma$ is to shift sharpness from $a$ to $a+\gamma$. So in the flat limit $a\ll 1/\eta$, a reasonable $\gamma<1/\eta - a$ can make $\sigma$ smaller, i.e., reducing valley variations.

\begin{figure}[ht]
    \centering
    \includegraphics[width=0.3\linewidth]{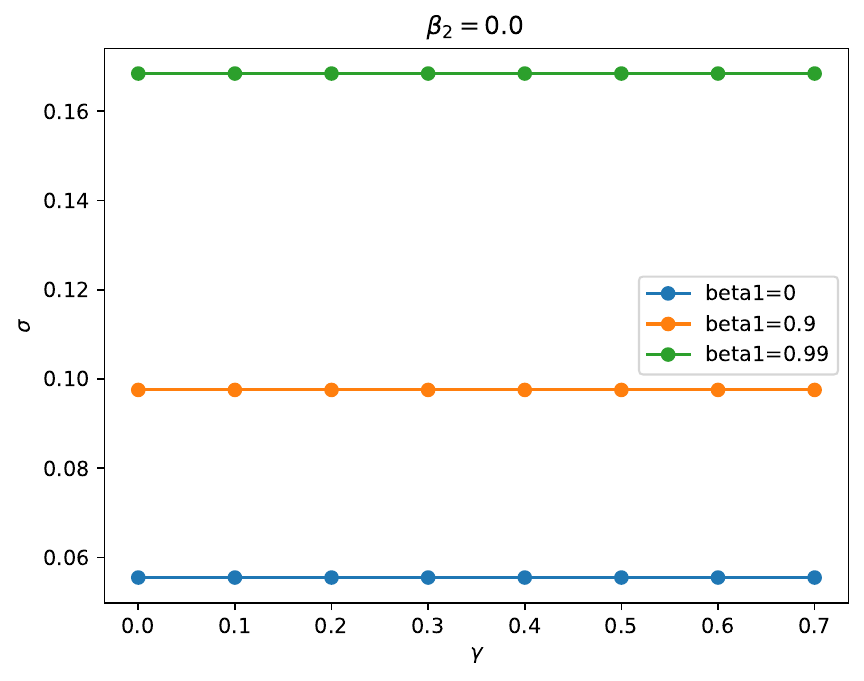}
    \includegraphics[width=0.3\linewidth]{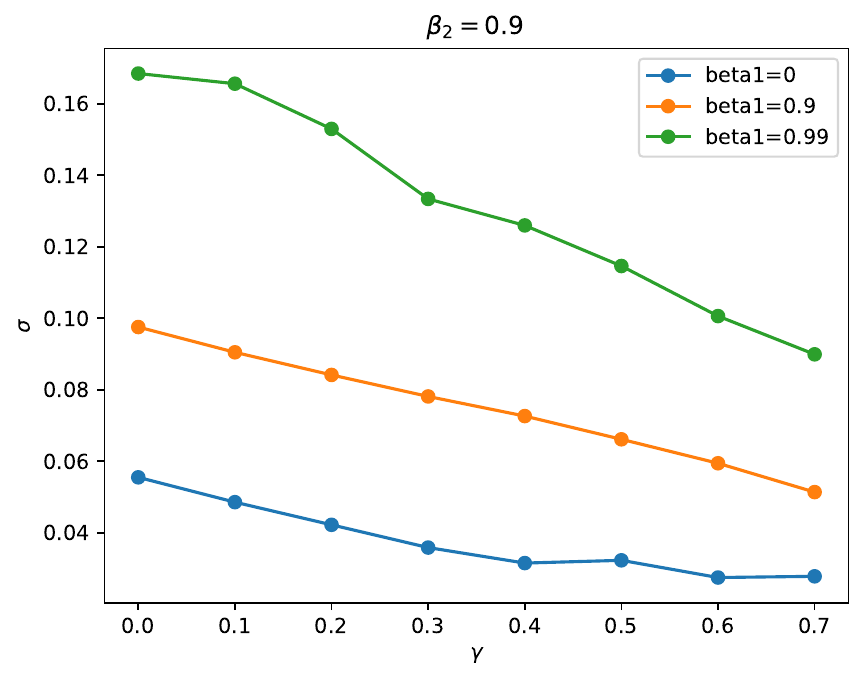}
    \includegraphics[width=0.3\linewidth]{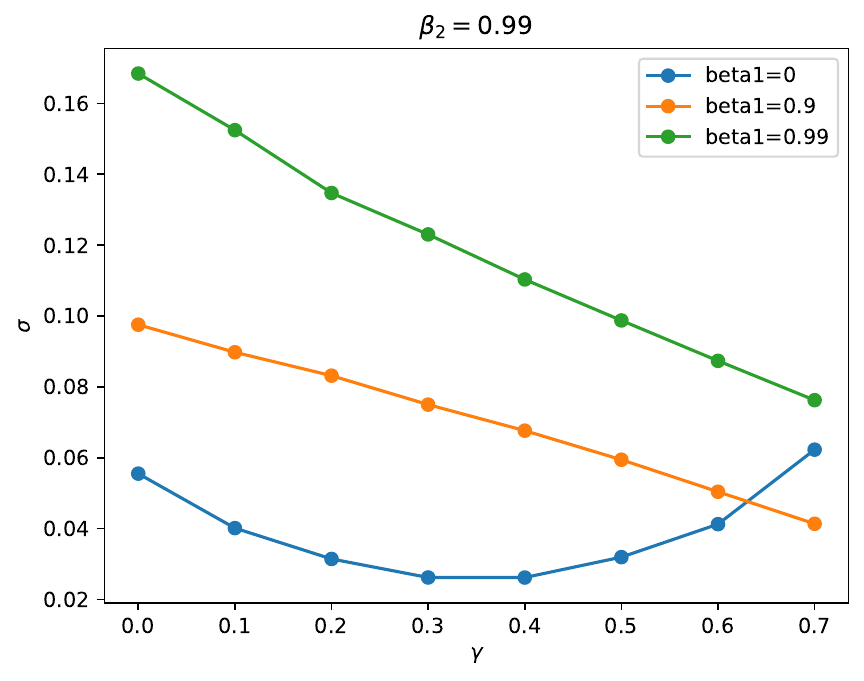}
    \caption{Gaussian width $\sigma$ for different self-attracting force $\gamma$, $\eta=0.1$, $a=1$, $\sigma_g=1$.}
    \label{fig:focus}
\end{figure}




\section{Designing learning rate schedules}\label{app:lr_schedule}

What insights can we gain to make training more effective? If we assume $\Delta_{\rm entropic}$ and $\Delta_{\rm anneal}$ can be ignored, the only way to reduce loss is by reducing $\ell(D,\eta_{\rm min})$, which involves reducing $\eta_{\rm min}$ and/or increasing $D$. 

However, reducing $\eta_{\rm min}$ may have a non-trivial effect on $\Delta_{\rm anneal}$, because in Section~\ref{sec:4} we have shown that the decay time $T_d\sim1/\eta_{\rm min}$, meaning that if one wants to reduce $\eta_{\rm min}$ by a factor of 2, the duration of the decay phase should be 2 times longer, which is not very efficient. 

\begin{figure}
    \centering
    \includegraphics[width=1.0\linewidth]{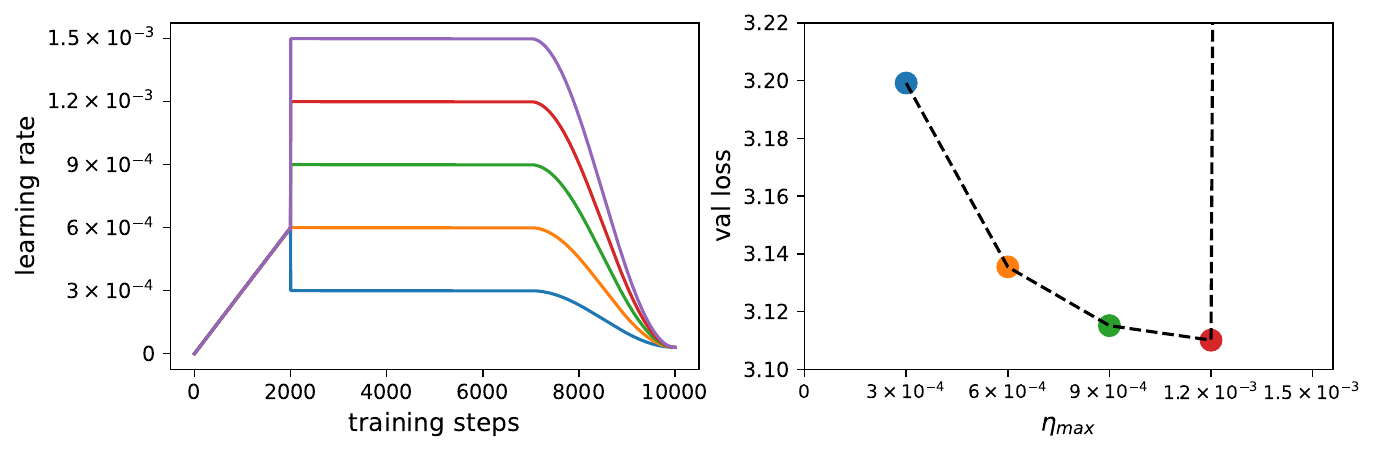}
    \caption{Learning rate schedules with different stable learning rate $\eta$. A larger $\eta$ leads to a lower validation loss, unless NaN issues occur (at $\eta=1.5\times 10^{-3}$).}
    \label{fig:eta_max}
\end{figure}

\begin{figure}
    \centering
    \includegraphics[width=1.0\linewidth]{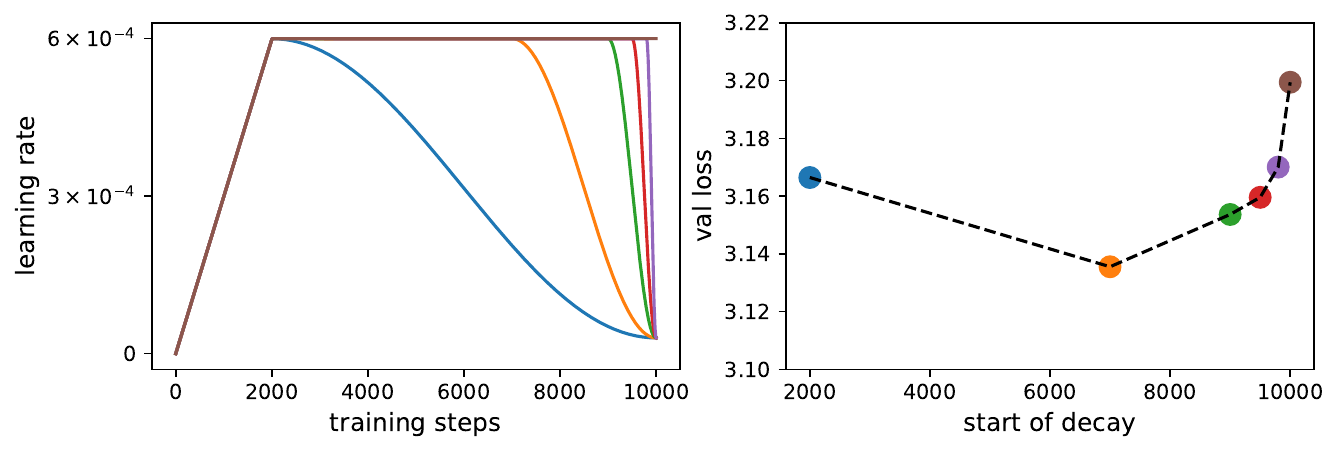}
    \caption{Learning rate schedules with different starting points of the decay phase. A longer stable phase does not necessarily lead to lower validation loss, since the decay phase cannot be too short due to annealing.}
    \label{fig:decay_start_iter}
\end{figure}

We now consider fixing $\eta_{\rm min}$ but increasing $D$ by choosing a larger $\eta$ in the stable phase. There are two potential concerns with this strategy: (1) Perhaps a longer decay schedule is needed to decay a larger $\eta$ to $\eta_{\rm min}$. Luckily, this is not a problem because Section~\ref{sec:4} showed that $T_d$ has an upper bound $O(1/\eta_{\rm min})$ which is independent of $\eta$. (2) Perhaps a larger $\eta$ includes larger entropic forces. Our $\eta$ sum alignment experiments in Section~\ref{sec:5} show that the effect of entropic forces is negligible. With both concerns cleared, our experiments in Figure~\ref{fig:eta_max} indeed show the efficiency of choosing a larger stable learning rate. However, the learning rate cannot be too large to cause numerical problems. Note that our experiments are done on two V-100s with float16 precision. More advanced machines are supposed to allow even higher learning rates without a blowup. We also note that our strategy is not just running the stable phase longer, which inevitably faces a trade-off: longer stable phase reduces $\ell(D,\eta_{\rm min})$ by increasing $D$, but potentially increases $\Delta_{\rm anneal}$ since the decay phase is eaten by the stable phase. The trade-off is shown in Figure~\ref{fig:decay_start_iter}.

\end{document}